\PassOptionsToPackage{table,svgnames}{xcolor}

\documentclass[10pt,twocolumn,letterpaper]{article}

\usepackage[pagenumbers]{cvpr} 

\usepackage{amsmath,amsfonts,bm}
\usepackage{amssymb}

\def\eqref#1{equation~\ref{#1}}

\def\1{\bm{1}}

\def\rvf{{\mathbf{f}}}

\def\rvu{{\mathbf{i}}}

\def\rvq{{\mathbf{q}}}

\def\rvu{{\mathbf{u}}}
\def\rvv{{\mathbf{v}}}

\def\rvx{{\mathbf{x}}}

\def\rvz{{\mathbf{z}}}

\def\rmA{{\mathbf{A}}}

\def\rmM{{\mathbf{M}}}

\DeclareMathAlphabet{\mathsfit}{\encodingdefault}{\sfdefault}{m}{sl}
\SetMathAlphabet{\mathsfit}{bold}{\encodingdefault}{\sfdefault}{bx}{n}

\usepackage{relsize}%

\usepackage{comment}
\usepackage{multirow}
\usepackage{pifont}
\usepackage{makecell}
\usepackage{booktabs,siunitx}
\usepackage[super]{nth}
\usepackage{soul}
\usepackage[most]{tcolorbox}

\usepackage{algorithm}
\usepackage{algorithmic}

\usepackage{arydshln}
\makeatletter
\def\adl@drawiv#1#2#3{%
	\hskip.5\tabcolsep
	\xleaders#3{#2.5\@tempdimb #1{1}#2.5\@tempdimb}%
	#2\z@ plus1fil minus1fil\relax
	\hskip.5\tabcolsep}
\newcommand{\cdashlinelr}[1]{%
	\noalign{\vskip\aboverulesep
		\global\let\@dashdrawstore\adl@draw
		\global\let\adl@draw\adl@drawiv}
	\cdashline{#1}
	\noalign{\global\let\adl@draw\@dashdrawstore
		\vskip\belowrulesep}}
\makeatother

\definecolor{Gray}{gray}{0.85}
\definecolor{whitesmoke}{rgb}{0.942, 0.942, 0.966}
\definecolor{LightCyan}{rgb}{0.88,1,1}
\newcolumntype{g}{>{\columncolor{whitesmoke}}c}

\newcommand{\xmark}{\ding{53}}

\definecolor{cvprblue}{rgb}{0.21,0.49,0.74}
\usepackage[pagebackref,breaklinks,colorlinks,allcolors=cvprblue]{hyperref}

\def\paperID{4936} 
\def\confName{CVPR}
\def\confYear{2026}

\title{Diffusion-Based Makeup Transfer with Facial Region-Aware Makeup Features}

\author{Zheng Gao$^1$\textsuperscript{*}, Debin Meng$^1$, Yunqi Miao$^2$, Zhensong Zhang$^2$, Songcen Xu$^2$, Ioannis Patras$^1$, Jifei Song$^2$ \\
	{\small $^1$Queen Mary University of London, $^2$Huawei London Research Center}
}

\begin{document}
\maketitle

\begin{abstract}
	Current diffusion-based makeup transfer methods commonly use the makeup information encoded by off-the-shelf foundation models (\eg, CLIP) as condition to preserve the makeup style of reference image in the generation. Although effective, these works mainly have two limitations: \textbf{(1)} foundation models pre-trained for generic tasks struggle to capture makeup styles; \textbf{(2)} the makeup features of reference image are injected to the diffusion denoising model as a whole for global makeup transfer, overlooking the facial region-aware makeup features (\ie, eyes, mouth, etc) and limiting the regional controllability for region-specific makeup transfer. To address these, in this work, we propose \textbf{F}acial \textbf{R}egion-\textbf{A}ware \textbf{M}akeup features (\textbf{FRAM}), which has two stages: \textbf{(1)} makeup CLIP fine-tuning; \textbf{(2)} identity and facial region-aware makeup injection. For makeup CLIP fine-tuning, unlike prior works using off-the-shelf CLIP, we synthesize annotated makeup style data using GPT-o3 and text-driven image editing model, and then use the data to train a makeup CLIP encoder through self-supervised and image-text contrastive learning. For identity and facial region-aware makeup injection, we construct before-and-after makeup image pairs from the edited images in stage 1 and then use them to learn to inject identity of source image and makeup of reference image to the diffusion denoising model for makeup transfer. Specifically, we use learnable tokens to query the makeup CLIP encoder to extract facial region-aware makeup features for makeup injection, which is learned via an attention loss to enable regional control. As for identity injection, we use a ControlNet Union to encode source image and its 3D mesh simultaneously. The experimental results verify the superiority of our regional controllability and our makeup transfer performance. Code is available at \href{https://github.com/zaczgao/Facial_Region-Aware_Makeup}{GitHub}.
\end{abstract}

\section{Introduction}
\begin{figure}[htb]
	\centering
	\includegraphics[width=\linewidth]{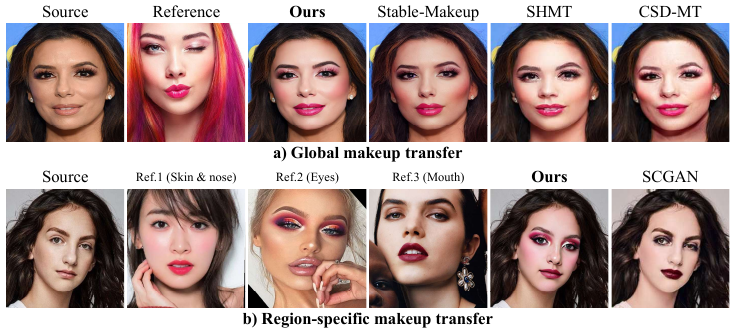}
	\caption{\textbf{Applications of our method \text{FRAM}}. Global makeup transfer copies the makeup style of reference image to source image. Region-specific makeup transfer combines the regional makeup styles (skin, eyes and mouth) of three reference images.}
	\label{fig:FRAM-app}
\end{figure}

Makeup transfer copies the makeup style from reference face image to source image, while preserving source face's identity~\cite{zhang2024stable}. With the success of large-scale text-to-image (T2I) diffusion~\cite{rombach2022high}, recent works condition T2I diffusion models on source identity image and reference makeup image to achieve high-fidelity makeup transfer~\cite{sun2024shmt,ruan2025mad,zhang2024stable,sii2024gorgeous}.

Current methods commonly inject the makeup features encoded by off-the-shelf foundation models (\eg, CLIP~\cite{radford2021learning}) to the diffusion denoising model for reference makeup preservation~\cite{sun2024shmt,zhang2024stable}. Despite their effectiveness, these methods are mainly limited in two aspects: \textbf{(1)} foundation models pre-trained on natural images are optimized for generic tasks, and therefore struggle to capture facial makeup styles; \textbf{(2)} these works inject the makeup features of reference image to the diffusion denoising model as a whole for global makeup transfer, overlooking the facial region-aware makeup features (\ie, eyes, mouth, etc) and limiting regional controllability for region-specific makeup transfer. To address these, first, unlike prior works that directly use off-the-shelf CLIP~\cite{radford2021learning}, we \textbf{train a makeup CLIP encoder} to encode makeup styles by synthesizing annotated makeup style data for CLIP fine-tuning. In~\cref{tab:FRAM-ablation-CLIP}, we show that our makeup CLIP encoder is a better makeup style encoder. Moreover, inspired by self-supervised facial representation learning~\cite{gao2024self}, we use learnable tokens as queries for the makeup CLIP encoder to extract \textbf{facial region-aware makeup features} for makeup injection. This is learned with an attention loss, which encourages the model to look at facial regions (see cross-attention maps in~\cref{fig:FRAM-region}) and enables region-specific makeup transfer (\cref{fig:FRAM-app}b and \cref{fig:FRAM-region}). This is in contrast to prior works that \textbf{inject makeup features as a whole} for global makeup transfer~\cite{sun2024shmt,zhang2024stable}.

In this work, we propose a diffusion-based facial makeup transfer framework, \textbf{F}acial \textbf{R}egion-\textbf{A}ware \textbf{M}akeup features (\textbf{FRAM}), which has two stages: \textbf{(1) makeup CLIP fine-tuning} learns a makeup CLIP encoder; \textbf{(2) identity and facial region-aware makeup injection} learns to extract identity features of source image with the ControlNet~\cite{zhang2023adding} and facial region-aware makeup features of reference image from our makeup CLIP, and then inject them to the diffusion denoising model for makeup transfer. For \textit{makeup CLIP fine-tuning}, as existing makeup datasets lack labels and text descriptions, we synthesize annotated makeup style data for CLIP fine-tuning by using GPT-o3 to generate makeup style descriptions and then prompt a state-of-the-art (SOTA) text-driven image editing model (\eg, FLUX.1-Kontext-dev~\cite{labs2025flux}) to add makeup to face images, obtaining images with various makeup styles. Inspired by CSD~\cite{somepalli2024measuring} that measures natural image style similarity, we fine-tune CLIP to learn content-invariant makeup features using 2 objectives: self-supervised and image-text contrastive learning. Unlike CSD~\cite{somepalli2024measuring} that only adopts image contrastive learning on natural artistic images, we further incorporate the text supervision of makeup style descriptions via image-text contrastive learning on synthetic face images to learn makeup features. For \textit{identity and facial region-aware makeup injection}, given the scarcity of paired data, we construct before-and-after makeup image pairs from the edited images in stage 1. The paired data is used to learn identity and makeup information injection for makeup transfer. We propose to extract facial region-aware makeup features to enable region-specific makeup transfer. Specifically, inspired by facial representation learning~\cite{gao2024self}, we use learnable tokens as queries for our makeup CLIP encoder to output a set of ``facial region makeup embeddings'', each associated with a facial region. These embeddings are treated as image prompts and injected to the denoising model through cross-attention to control the sampling process. To supervise the learning of these embeddings, we align the cross-attention maps of these image prompts and the face parsing masks (produced by a face parsing model FaRL~\cite{zheng2022general}) of reference image (\ie, synthesized makeup image of source, which shares the same masks as source). This enables the learned queries to guide the model to apply makeup to source's facial regions \textbf{without using masks} during inference. As for identity injection, unlike Stable-Makeup~\cite{zhang2024stable} that uses two ControlNets~\cite{zhang2023adding}, we use a ControlNet Union~\cite{ControlNet++} to leverage the pixel-level identity information of source image and the structure guidance from 3D mesh~\cite{wang20243d} simultaneously.

Our contributions can be summarized as follows:
\begin{itemize}
	\item We propose a diffusion-based makeup transfer framework, \textbf{F}acial \textbf{R}egion-\textbf{A}ware \textbf{M}akeup features (\textbf{FRAM}).
	\item We learn a makeup CLIP encoder by fine-tuning CLIP on synthesized annotated makeup style data.
	\item We construct before-and-after makeup image pairs to learn makeup transfer. Moreover, we make a first attempt to enable diffusion-based region-specific makeup transfer by using learnable queries to extract facial region-aware makeup features from our makeup CLIP encoder.
\end{itemize}

\begin{figure*}[htb]
	\centering
	\includegraphics[width=0.95\linewidth]{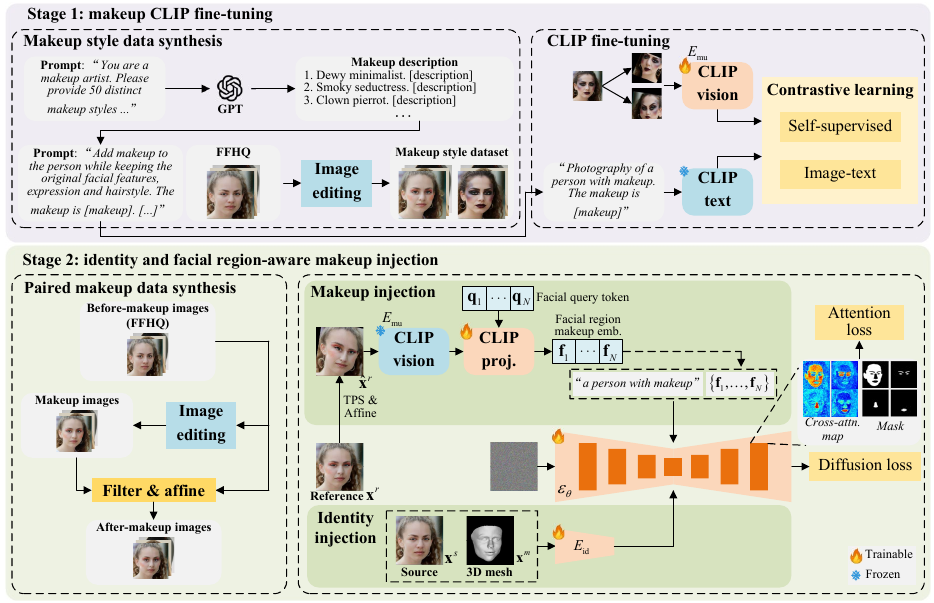}
	\caption{\textbf{Overview of our FRAM}. It has two stages: \textbf{(1)} makeup CLIP fine-tuning; \textbf{(2)} identity and facial region-aware makeup injection. For stage 1, we synthesize annotated makeup style data using GPT-o3 and a SOTA text-driven image editing model (\eg, FLUX.1-Kontext-dev~\cite{labs2025flux}), and then use it to learn a makeup CLIP vision encoder $E_\text{mu}$. For stage 2, we construct before-and-after makeup image pairs from the edited images in stage 1 and use them to learn to transfer the makeup style of reference image $\rvx^r$ to source image $\rvx^s$, \ie, inject identity and makeup information to the diffusion denoising model $\epsilon_\theta$. For makeup injection, we extract facial region makeup embeddings $\{ \rvf_n \}^N_{n=1}$ by querying our makeup CLIP encoder $E_\text{mu}$ and inject them to $\epsilon_\theta$ via cross-attention. For identity injection, we adopt a ControlNet Union~\cite{ControlNet++} to encode pixel-level identity information of source image $\rvx^s$ and structure guidance from 3D mesh $\rvx^m$ simultaneously.}
	\label{fig:FRAM-overview}
\end{figure*}

\begin{figure}[h]
	\centering
	\includegraphics[width=0.95\linewidth]{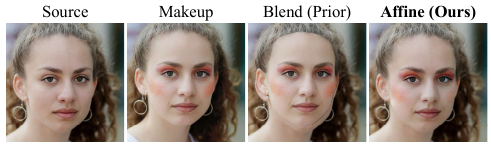}
	\caption{\textbf{Affine \vs blend face and non-face region}.}
	\label{fig:FRAM-affine}
\end{figure}

\section{Related work}
\subsection{Facial makeup transfer}
Makeup transfer based on GAN (Generative Adversarial Network~\cite{goodfellow2014generative}) faces challenges in transferring complex real-world makeup styles~\cite{li2018beautygan,jiang2020psgan,nguyen2021lipstick,deng2021spatially,xiang2022ramgan,yan2023beautyrec,sun2024content,lu2025beautybank,jin2024toward}. To address this, recent works leverage pre-trained diffusion model~\cite{feng2026noisy,meng2024mm2latent,meng2025training} for makeup transfer by using source identity image and reference makeup image as conditions, which are injected to the diffusion denoising model to control the sampling process~\cite{sun2024shmt,ruan2025mad,zhang2024stable,sii2024gorgeous}. Stable-Makeup~\cite{zhang2024stable} uses the ControlNet~\cite{zhang2023adding} and off-the-shelf CLIP~\cite{radford2021learning} to extract identity and makeup features, respectively. In contrast, we synthesize annotated makeup style data and fine-tune CLIP on it to learn a makeup CLIP encoder. Moreover, unlike Stable-Makeup~\cite{zhang2024stable} that injects CLIP features to the diffusion denoising model as a whole for global makeup transfer, we learn to extract facial region-aware makeup features from our makeup CLIP, enabling region-specific makeup transfer. Gorgeous~\cite{sii2024gorgeous} encodes makeup styles by learning text embeddings via textual inversion~\cite{gal2023an}, which are used as the text condition for denoising model. In contrast, we use our makeup CLIP encoder as the makeup encoder instead of optimizing text embeddings, which eliminates the need to learn a new token for each new concept of makeup style. SHMT~\cite{sun2024shmt} proposes a self-supervised method to learn the identity and makeup injection using existing makeup datasets. By contrast, we synthesize before-and-after makeup image pairs for supervised learning. MAD~\cite{ruan2025mad} is a training-free method that blends source and reference image to obtain the noise seed so that the makeup and identity information are encoded during sampling. However, MAD~\cite{ruan2025mad} is less effective than the training-based methods above. Concurrent works~\cite{zhu2025flux,wu2025evomakeup} also apply image editing to generate paired makeup data. However, these works simply discard the generated images with large face region misalignment while we alleviate this by aligning the face regions via affine transformation (\cref{fig:FRAM-affine}). We additionally show that the aligned data leads to better performance in the supplementary. Moreover, these works lack regional controllability of several references.

\subsection{Paired makeup data generation}
Recent works use a training-free image editing model to add makeup to face images to obtain before-and-after makeup image pairs~\cite{zhang2024stable,lu2025beautybank}. However, the quality of generated makeup image is limited by the training-free model. In contrast, we use the SOTA training-based image editing model (\eg, FLUX.1-Kontext-dev~\cite{labs2025flux}) to edit face images. Moreover, to avoid edits on non-face region, these works commonly combine the face region of generated makeup image and non-face region of source identity image to obtain the before-and-after makeup pair, under the guidance of face parsing masks. However, the face region of generated makeup image may not be spatially aligned with the face region of source image. Thus, the composite image is distorted. To alleviate this, we align face regions between the generated makeup image and source image via affine transformation. Another work, TinyBeauty~\cite{jin2024toward} uses 5 predefined makeup styles from commercial photo editor MEITU to fine-tune the diffusion model for paired data generation. FFHQ-Makeup~\cite{yang2025ffhq} synthesizes paired data via makeup transfer on makeup styles from existing datasets. However, the diversity of makeup styles in these works is limited. In contrast, we use GPT-o3 to generate various makeup style descriptions and then prompt an image editing model to synthesize face images with diverse makeup styles. Unlike the above methods that add makeup to face, some~\cite{gu2019ladn,sun2022ssat,sii2024gorgeous} remove the makeup of face to obtain paired data. However, these works are suboptimal due to the limited makeup style diversity of existing makeup datasets.

\subsection{Facial region discovery}
GAN-based methods have also explored facial region discovery for makeup transfer~\cite{deng2021spatially,xiang2022ramgan}. SCGAN~\cite{deng2021spatially} extracts makeup features for facial regions at the image level, under the guidance of face parsing masks. RamGAN~\cite{xiang2022ramgan} extracts makeup features for facial regions by aggregating feature maps at the feature level. Despite different techniques, these methods are tailored for GAN while we propose to query our makeup CLIP encoder and inject facial region-aware makeup features to the diffusion denoising model.

\section{Methodology}
\subsection{Overview}\label{sec:FRAM-overview}
An overview of our FRAM is shown in~\cref{fig:FRAM-overview}. Given a reference makeup image $\rvx^r$ and source identity image $\rvx^s$, the goal is to leverage the diffusion denoising process to transfer the makeup style of $\rvx^r$ to $\rvx^s$, while preserving the identity of $\rvx^s$. Following common practice~\cite{rombach2022high}, the diffusion denoising model $\epsilon_\theta$ operates in the latent space of a Variational Autoencoder (VAE)~\cite{esser2021taming}. During training, Gaussian noise $\epsilon$ is added to $\rvx^r$ to obtain noised latent $\rvx_t$ at timestep $t \in \{0,\ldots,T\}$, where $\rvx_0 = \rvx^r$. $\epsilon_\theta$ is trained to predict the added noise to denoise $\rvx_t$. During sampling, $\epsilon_\theta$ recovers the clean sample $\rvx_0$ by gradually denoising the Gaussian noise via the denoising process. To control the denoising, the identity features of $\rvx^s$ and makeup features of $\rvx^r$ are injected to $\epsilon_\theta$ as conditions. The training of our FRAM has two stages: \textbf{(1)} makeup CLIP fine-tuning; \textbf{(2)} identity and facial region-aware makeup injection. For stage 1, we synthesize annotated makeup style data and use it to learn a makeup CLIP encoder $E_\text{mu}$ (\cref{sec:FRAM-clip}). For stage 2, we construct before-and-after makeup image pairs and use them to learn makeup transfer by injecting identity and makeup information to $\epsilon_\theta$, \ie, identity features from an identity encoder $E_\text{id}$ and facial region-aware makeup features from our makeup CLIP encoder $E_\text{mu}$ (\cref{sec:FRAM-inject}).

\subsection{Stage 1: makeup CLIP fine-tuning}\label{sec:FRAM-clip}
\textbf{Makeup style data synthesis}. To synthesize annotated makeup style data with various makeup styles for CLIP fine-tuning, we prompt GPT-o3 to generate 50 distinct, diverse and detailed makeup style descriptions, covering styles from light to heavy makeup, \eg, ``\textit{Dewy minimalist. [description]}'', where ``[description]'' denotes the detailed description of makeup style ``dewy minimalist''. Then we use these makeup styles to prompt a SOTA text-driven image editing model (\eg, FLUX.1-Kontext-dev~\cite{labs2025flux}) to add makeup to face images from the dataset FFHQ~\cite{karras2019style} to obtain the makeup style images. The prompt template is ``\textit{Add makeup to the person while keeping the original facial features, expression and hairstyle. The makeup is [makeup]. Maintain the original position, background, camera angle, framing, and perspective.}'', where ``[makeup]'' is the makeup style description from GPT-o3. The SOTA image editing model ensures the facial identity is preserved in the generation while FFHQ ensures facial diversity. More details and samples of generated makeup style descriptions and images are provided in the supplementary.

\textbf{CLIP fine-tuning}. Next, we fine-tune a CLIP vision encoder to learn a makeup encoder $E_\text{mu}$, using self-supervised and image-text contrastive learning. Given a face image $\rvx_i$ from the synthesized makeup style data, we augment its content to obtain two views of $\rvx_i$, \ie, $\{\rvx_{i,v} \mid v \in \{1,2\}\}$, which exhibit altered facial content while preserving the makeup style. Following~\cite{somepalli2024measuring,zhang2024stable}, we use Thin Plate Spline (TPS)~\cite{wahba1990spline}, random crop, flip, and affine transformation for augmentations. Each $\rvx_{i,v}$ is fed into $E_\text{mu}$ to obtain the embedding, \ie, $\rvz_{i,v} = E_\text{mu}(\rvx_{i,v})$. Let $I = \{ (i,v) \mid i \in \{1,\ldots,B\}, v \in \{1,2\} \}$ be the set of indices of samples in the mini-batch. We use self-supervised contrastive learning (SSL) to fine-tune CLIP to maximize the similarity of the two views, which encourages CLIP to ignore face structure while preserving high-level semantics (\eg, makeup style). The InfoNCE loss~\cite{pmlr-v119-chen20j} is adopted for SSL:
\begin{equation}\label{eq:FRAM-ssl}
	\mathcal{L}_\text{ssl} = -\log{\frac{\exp{(f_{s}{(\rvz_{i,1},\rvz_{i,2})/\tau)}}} {\sum_{a \in I(i,1)}{\exp{(f_{s}{(\rvz_{i,1}, \rvz_a)/\tau})}}}},
\end{equation}
where $f_{s}(\rvu, \rvv)=\frac{\rvu^\top\rvv}{{\lVert\rvu\rVert}_2{\lVert\rvv\rVert}_2}$ denotes the cosine similarity between $\rvu$ and $\rvv$, $\tau$ is the temperature, and $I(i,1) \equiv I \setminus \{(i,1)\}$. The above objective only uses the image embeddings. We further align the image embedding of $E_\text{mu}$ with the text embedding of a pre-trained CLIP text encoder, which encodes the makeup style descriptions. The CLIP text encoder takes as input the prompt template ``\textit{Photography of a person with makeup. The makeup is [makeup]}'', where ``[makeup]'' is the makeup style description. The image-text contrastive learning objective is:
\begin{equation}\label{eq:FRAM-text}
	\mathcal{L}_\text{text} = 
	-\frac{1}{|P|} \sum_{p \in P} {\log{\frac{\exp{(f_{s}{(\rvz_{i,1},\rvz^\text{text}_p)/\tau)}}}{\sum_{a \in I}{\exp{(f_{s}{(\rvz_{i,1}, \rvz^\text{text}_a)/\tau})}}}}},
\end{equation}
where $P$ is the set of indices of positive samples (same makeup style as $\rvz_{i,1}$) in the mini-batch, $|P|$ is the number of positive samples, $\rvz^\text{text}_p$ and $\rvz^\text{text}_a$ are text embeddings for positive sample of $\rvz_{i,1}$ and sample in the batch, respectively.

The overall CLIP fine-tuning objective is as follows:
\begin{equation}\label{eq:FRAM-pre}
	\mathcal{L}_\text{clip} = \mathcal{L}_\text{ssl} + \mathcal{L}_\text{text},
\end{equation}

We initialize $E_\text{mu}$ with pre-trained CLIP and fine-tune the last encoder layer. The learned $E_\text{mu}$ is used to extract the makeup features of reference image in stage 2 (\cref{sec:FRAM-inject}).

\subsection{Stage 2: identity and facial region-aware makeup injection}\label{sec:FRAM-inject}
\textbf{Paired makeup data synthesis}. To construct before-and-after makeup image pairs, we conduct filtering and alignment on the pairs of FFHQ~\cite{karras2019style} and makeup style images in~\cref{sec:FRAM-clip}. First, we use GPT-5 to filter out FFHQ-makeup image pairs with non-realistic makeup images or different facial identities and expressions. To avoid editing non-face region during image editing, prior works commonly blend the face region of generated makeup image and non-face region of source identity image to obtain the before-and-after makeup pair, under the guidance of face parsing masks~\cite{zhang2024stable,lu2025beautybank}. However, as shown in~\cref{fig:FRAM-affine}, image editing could cause spatial misalignment, \ie, the face region of generated makeup image (\nth{2} column) may not be spatially aligned with the face region of source image. Thus, the composite image (\nth{3} column) is distorted. To mitigate this, we perform affine transformation based on the facial landmarks (predicted by JMLR~\cite{guo2022perspective}) to replace source image's face with that of generated makeup image. This aligns the face regions between the generated image and source image to obtain the after-makeup image (\nth{4} column). Next, the images with teeth and eye misalignments relative to the source image are filtered out based on the IoU (Intersection over Union) of their face parsing masks (produced by FaRL~\cite{zheng2022general}). More details are in the supplementary.

With the synthesized image pairs, we learn to inject identity and makeup information to $\epsilon_\theta$ for makeup transfer.

\textbf{Makeup injection}. Inspired by face representation learning, we extract facial region-aware makeup features from $E_\text{mu}$. First, we follow~\cite{zhang2024stable} to augment the structure of $\rvx^r$ via transformations (TPS and affine) to obtain $\hat{\rvx}^r$. Then, a CLIP projector takes as input $N$ learnable tokens $\{ \rvq_n \}^N_{n=1}$ (query) and the CLIP features from the last encoder layer $E_\text{mu}(\hat{\rvx}^r)$ (key and value) to predict $N$ ``facial region makeup embeddings'' $\{ \rvf_n \}^N_{n=1}$, each associated with a facial region. The CLIP projector adopts the structure used in InstantID~\cite{wang2024instantid}. Similar to IP-Adapter~\cite{ye2023ip}, the embeddings $\{ \rvf_n \}^N_{n=1}$ are treated as \textbf{image prompt} embeddings, allowing $\epsilon_\theta$ to inject the makeup features via cross-attention. The text prompt for $\epsilon_\theta$ is ``\textit{a person with makeup}''. Let $\rmA^l_n$ be the cross-attention map of $\rvf_n$ (facial region $n$) at layer $l$, $\bar{\rmA}_n = \frac{1}{L}\sum_{l=1}^{L}\rmA^l_n$ be the averaged cross-attention map across $L$ layers. To supervise the learning of $\{ \rvf_n \}^N_{n=1}$, we align the cross-attention maps of $\{ \rvf_n \}^N_{n=1}$ and face parsing masks of $\rvx_r$ using the following attention loss $\mathcal{L}_\text{attn}$:
\begin{align}\label{eq:FRAM-attn}
	\mathcal{L}_\text{attn} & = \frac{1}{NUV} \sum_{n=1}^{N} \sum_{u,v}
	\Big[ \text{FL}(\bar{\rmA}_n[u,v], \rmM_n[u,v]) \nonumber \\
	&\quad + \text{DICE}(\bar{\rmA}_n[u,v], \rmM_n[u,v]) \Big],
\end{align}
where $U$ and $V$ are the attention map size, $[u,v]$ is the spatial location, $\rmM_n$ is the binary mask for facial region $n$ (\ie, skin, eyes, nose, and mouth) predicted by a face parsing model~\cite{zheng2022general}. $\mathcal{L}_\text{attn}$ encourages $\{ \rvf_n \}^N_{n=1}$ to look at the corresponding facial regions in the cross-attention maps. $\mathcal{L}_\text{attn}$ adopts the binary mask loss in segmentation~\cite{carion2020end}, which consists of a focal loss~\cite{lin2017focal} $\text{FL}()$ and a dice loss~\cite{milletari2016v} $\text{DICE}()$. As shown in~\cref{sec:FRAM-region}, the learned queries enable regional control without relying on masks during inference.

\textbf{Identity injection}. We use a ControlNet Union~\cite{ControlNet++} as the identity encoder $E_\text{id}$ for identity injection. $E_\text{id}$ encodes the pixel-level information from source image $\rvx^s$ and facial structure from 3D mesh $\rvx^m$ in a single ControlNet. We reconstruct 3D mesh $\rvx^m$ from $\rvx^s$ using  3DDFA-V3~\cite{wang20243d}. Null text is used as the prompt for $E_\text{id}$.

\textbf{Overall objective}. In addition to $\mathcal{L}_\text{attn}$ (\cref{eq:FRAM-attn}), we adopt the diffusion loss~\cite{ho2020denoising} to learn the joint injection of identity and makeup during the denoising process: 
\begin{equation}\label{eq:FRAM-diff}
	\mathcal{L}_\text{diff} = \mathbb{E}_{\rvx_0, t, \epsilon}\left\| \epsilon - \epsilon_\theta(\rvx_t,t,C) \right\|_2^2,
\end{equation}
where $C$ is the conditions, including prompt, identity and makeup features. The overall objective is as follows:
\begin{equation}\label{eq:FRAM-obj}
	\mathcal{L} = \mathcal{L}_\text{diff} + \mathcal{L}_\text{attn}.
\end{equation}

Since the makeup features are injected via cross-attention, we apply LoRA (Low-Rank Adaptation)~\cite{hu2022lora} to the cross-attention layers of pre-trained denoising model $\epsilon_\theta$ for fine-tuning. The LoRA layers of $\epsilon_\theta$, CLIP projector and identity encoder $E_\text{id}$ are jointly updated during training.


\subsection{Regional control}\label{sec:FRAM-region}
\begin{figure*}[htb]
	\centering
	\includegraphics[width=\linewidth]{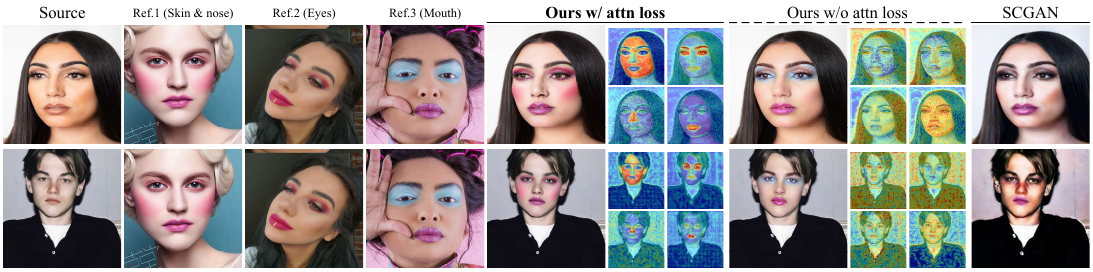}
	\caption{\textbf{Region-specific makeup transfer and qualitative ablation on attention loss}. The visualization of the cross-attention maps corresponding to the facial region makeup embeddings is next to the generated image.}
	\label{fig:FRAM-region}
\end{figure*}

Since the facial region makeup embeddings $\{ \rvf_n \}^N_{n=1}$ correspond to distinct facial regions, regional control can be achieved during sampling while prior diffusion-based methods lack such controllability. In~\cref{fig:FRAM-region}, we construct a new set of embeddings by selecting the region-specific embeddings from different reference images (\ie, skin from reference 1, eyes from reference 2 and mouth from reference 3), which are injected to the denoising model for region-specific makeup transfer.

\setlength{\tabcolsep}{1pt}
\begin{table*}[htb]
	\caption{\textbf{Quantitative results}. The \textbf{\textcolor{red}{red}}/\textbf{\textcolor{blue}{blue}} value indicates the top/second-ranked result, respectively.}
	\centering
	\resizebox{\linewidth}{!}{
		\begin{tabular}{l c c c c c c c c c c c c c c c}
			\toprule
			\multirow{2}{*}{Method} & \multicolumn{5}{c}{MT~\cite{li2018beautygan}} & \multicolumn{5}{c}{Wild-MT~\cite{jiang2020psgan}} & \multicolumn{5}{c}{CPM-real~\cite{nguyen2021lipstick}} \\
			\cmidrule(lr){2-6} \cmidrule(lr){7-11} \cmidrule(lr){12-16}
			& CSD $\uparrow$ & ID $\uparrow$ & SSIM $\uparrow$ & L2-M $\downarrow$ & Aes $\uparrow$ & CSD $\uparrow$ & ID $\uparrow$ & SSIM $\uparrow$ & L2-M $\downarrow$ & Aes $\uparrow$ & CSD $\uparrow$ & ID $\uparrow$ & SSIM $\uparrow$ & L2-M $\downarrow$ & Aes $\uparrow$ \\
			\midrule
			\multicolumn{9}{l}{\textbf{GAN-based}} \\
			CSD-MT~\cite{sun2024content} & 0.434 & 0.585 & 0.424 & 0.146 & 4.50 & 0.428 & 0.585 & 0.430 & 0.124 & 4.41 & 0.418 & 0.522 & 0.410 & 0.128 & 4.41 \\
			\midrule
			\multicolumn{9}{l}{\textbf{Diffusion-based}} \\
			MAD~\cite{ruan2025mad} & 0.328 & 0.535 & 0.805 & 0.003 & 4.26 & 0.401 & \textbf{\textcolor{blue}{0.541}} & 0.808 & 0.002 & 4.18 & 0.317 & \textbf{\textcolor{blue}{0.491}} & 0.788 & 0.003 & 4.18 \\
			Gorgeous~\cite{sii2024gorgeous} & 0.417 & \textbf{\textcolor{red}{0.652}} & \textbf{\textcolor{red}{0.896}} & 0.003 & 4.70 & 0.472 & \textbf{\textcolor{red}{0.628}} & \textbf{\textcolor{red}{0.887}} & 0.002 & 4.64 & 0.399 & \textbf{\textcolor{red}{0.624}} & \textbf{\textcolor{red}{0.871}} & \textbf{\textcolor{red}{0.002}} & 4.61 \\
			SHMT~\cite{sun2024shmt} & 0.498 & 0.372 & 0.811 & 0.012 & 4.86 & 0.516 & 0.411 & 0.820 & 0.010 & 4.78 & 0.417 & 0.408 & 0.793 & 0.013 & 4.64 \\
			Stable-M~\cite{zhang2024stable} & \textbf{\textcolor{blue}{0.527}} & 0.413 & 0.864 & 0.006 & \textbf{\textcolor{blue}{5.10}} & \textbf{\textcolor{blue}{0.562}} & 0.428 & 0.864 & \textbf{\textcolor{blue}{0.006}} & \textbf{\textcolor{blue}{5.02}} & \textbf{\textcolor{blue}{0.515}} & 0.415 & 0.777 & 0.007 & \textbf{\textcolor{blue}{4.87}} \\
			\rowcolor{whitesmoke} FRAM (\textbf{Ours}) & \textbf{\textcolor{red}{0.536}} & \textbf{\textcolor{blue}{0.587}} & \textbf{\textcolor{blue}{0.880}} & \textbf{\textcolor{red}{0.002}} & \textbf{\textcolor{red}{5.25}} & \textbf{\textcolor{red}{0.571}} & 0.499 & \textbf{\textcolor{blue}{0.866}} & \textbf{\textcolor{red}{0.002}} & \textbf{\textcolor{red}{5.05}} & \textbf{\textcolor{red}{0.528}} & 0.429 & \textbf{\textcolor{blue}{0.797}} & \textbf{\textcolor{blue}{0.003}} & \textbf{\textcolor{red}{4.95}} \\
			\bottomrule
		\end{tabular}
	}
	\label{tab:FRAM-result}
\end{table*}
\setlength{\tabcolsep}{6pt}

\begin{figure*}[htb]
	\centering
	\includegraphics[width=0.95\linewidth]{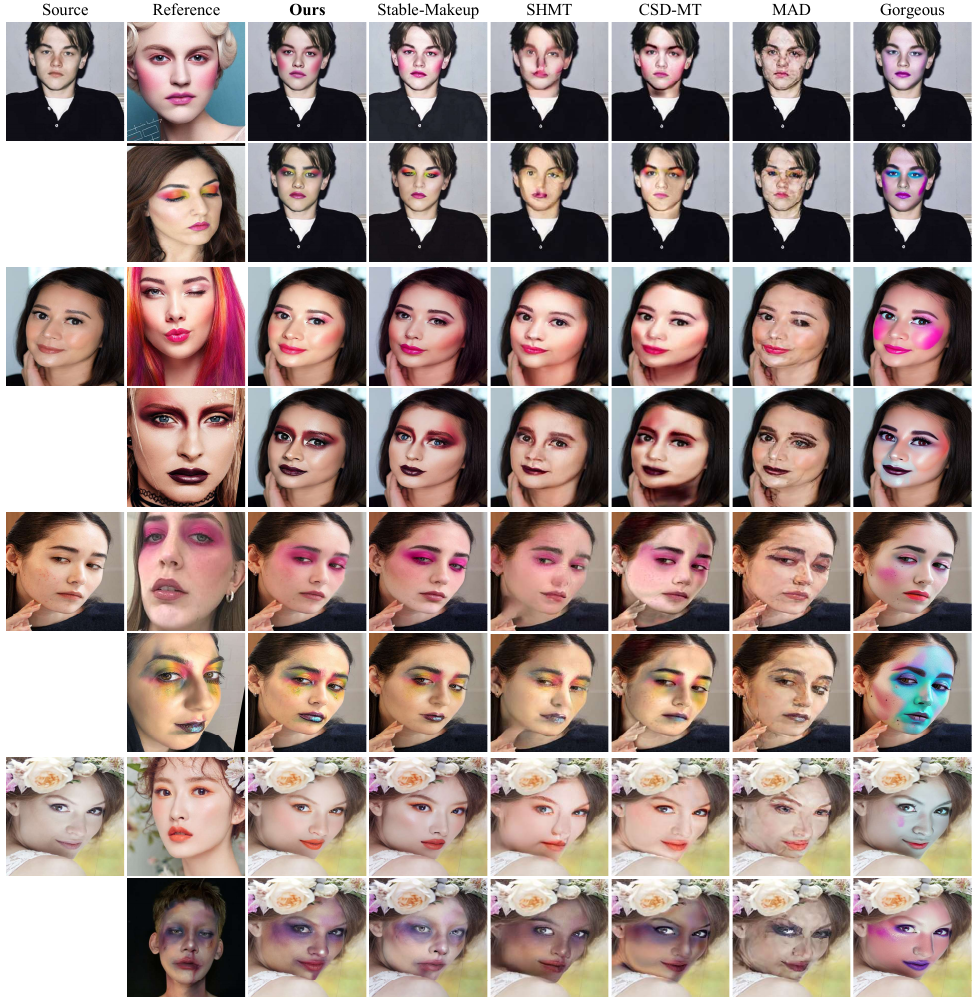}
	\caption{\textbf{Qualitative results}.}
	\label{fig:FRAM-result}
\end{figure*}

\begin{figure}[htb]
	\centering
	\includegraphics[width=\linewidth]{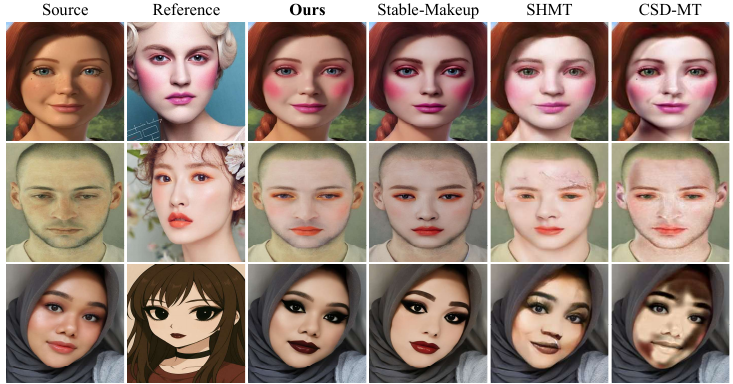}
	\caption{\textbf{Qualitative results for cross-domain faces}. Note that the reference in the last row is generated by GPT-Image-1.}
	\label{fig:FRAM-domain}
\end{figure}

\begin{figure}[htb]
	\centering
	\begin{subfigure}[b]{0.49\textwidth}
		\centering
		\includegraphics[width=\linewidth]{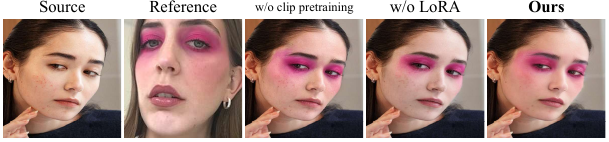}
		\caption{Makeup injection.}
		\label{fig:FRAM-ablation-makeup}
	\end{subfigure}
	\vfill
	\begin{subfigure}[b]{0.49\textwidth}
		\centering
		\includegraphics[width=\linewidth]{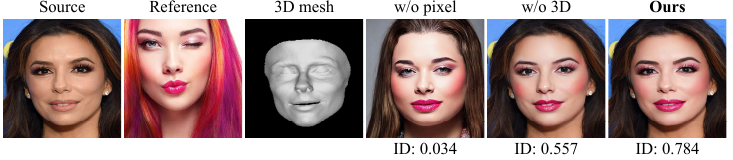}
		\caption{Identity injection. The ID score between the source image and generated image is reported below the generation.}
		\label{fig:FRAM-ablation-id}
	\end{subfigure}
	\caption{\textbf{Qualitative ablation on makeup/identity injection}.}
	\label{fig:FRAM-ablation-module}
\end{figure}

\section{Experiments}

\subsection{Experimental setups}
\textbf{Implementation}. The number of facial regions $N$ is set to 4, corresponding to skin, eyes, nose, and mouth. The CLIP makeup encoder $E_\text{mu}$ adopts ViT-L/14~\cite{dosovitskiy2021an}. Following~\cite{sii2024gorgeous}, we only edit the face region via an image inpainting pipeline to avoid edits on non-face area. We use Stable Diffusion v2.1 (SDv2.1)~\cite{rombach2022high} as the pre-trained diffusion model for all methods when possible and baseline's default model otherwise. For Stable-Makeup~\cite{zhang2024stable}, we retrain it with SDv2.1. The baselines adopt the default parameters. Following~\cite{zhang2024stable}, we evaluate on three datasets: Makeup Transfer (MT)~\cite{li2018beautygan}, Wild-MT~\cite{jiang2020psgan} and CPM-real~\cite{nguyen2021lipstick}. We perform ablation study on CPM-real~\cite{nguyen2021lipstick}. The image is resized to $512 \times 512$. More details and results can be found in the supplementary.

\textbf{Baselines}. We compare with GAN-based model (\ie, CSD-MT~\cite{sun2024content}) and diffusion-based models (\ie, SHMT~\cite{sun2024shmt}, MAD~\cite{ruan2025mad}, Gorgeous~\cite{sii2024gorgeous}, and Stable-Makeup~\cite{zhang2024stable} (Stable-M)). Since prior diffusion-based models don't support regional control, we compare with GAN-based SCGAN~\cite{deng2021spatially} for region-specific makeup transfer.

\textbf{Metrics}. Following~\cite{zhang2024stable,sii2024gorgeous,lu2025beautybank}, we use 5 metrics for the quantitative results: \textbf{(1)} CSD~\cite{somepalli2024measuring} measures the makeup style similarity between the generated image and reference makeup image, using the cosine similarity of a pre-trained style encoder. \textbf{(2)} ID measures the facial identity preservation between the generated image and source identity image, using the cosine similarity of a face identity encoder BlendFace~\cite{shiohara2023blendface}, which is robust to makeup. \textbf{(3)} SSIM~\cite{wang2004image} measures the face structure similarity between the generation and source. \textbf{(4)} L2-M~\cite{zhang2024stable} measures the non-edited region difference by calculating the MSE of non-face region between the generated image and source identity image. \textbf{(5)} Aesthetic Score~\cite{schuhmann2022laionb} (Aes) measures the quality of the generated image using the LAION-Aesthetics Predictor.

\subsection{Qualitative results}
In~\cref{fig:FRAM-result}, we report the qualitative results on global makeup transfer. We observe: \textbf{(1)} GAN-based CSD-MT~\cite{sun2024content} faces challenges in generating high-quality images that capture makeup styles. \textbf{(2)} The training-free method MAD~\cite{ruan2025mad} also struggles with image quality and makeup consistency. \textbf{(3)} Gorgeous~\cite{sii2024gorgeous} fails to capture makeup styles. \textbf{(4)} Stable-Makeup~\cite{zhang2024stable} has inferior results in preserving identity (\eg, \nth{7} row) and capturing the fine-grained makeup patterns (\eg, highlighted effect around the eyes in \nth{4} row).

In~\cref{fig:FRAM-domain}, we additionally provide results on cross-domain faces (\eg, cartoon, painting), demonstrating our superiority against SOTA methods.

\subsection{Quantitative results}
In~\cref{tab:FRAM-result}, the quantitative results are reported. We observe: \textbf{(1)} Although MAD~\cite{ruan2025mad} and Gorgeous~\cite{sii2024gorgeous} exhibit strong identity preservation (high ID score), both struggle to capture the desired makeup style (low CSD score). Moreover, MAD~\cite{ruan2025mad} suffers from low image quality (low Aes score). This is aligned with the observation in~\cref{fig:FRAM-result}. \textbf{(2)} Although Stable-Makeup~\cite{zhang2024stable} has strong makeup consistency results, it's inferior with identity preservation, which is also observed in~\cref{fig:FRAM-result}. \textbf{(3)} Compared with prior makeup transfer methods, our FRAM has a better balance between identity preservation and makeup consistency without sacrificing the image quality.

\setlength{\tabcolsep}{3pt}
\begin{table}[htb]
	\caption{\textbf{User study}. The best selection ratio ($\%$) is reported.}
	\centering
	\begin{tabular}{c c c g}
		\toprule
		CSD-MT & SHMT & Stable-Makeup & Ours \\
		\midrule
		20.5\% & 5\% & 16\% & 58.5\% \\
		\bottomrule
	\end{tabular}
	\label{tab:FRAM-user_study}
\end{table}
\setlength{\tabcolsep}{6pt}

\subsection{User study}
Following~\cite{zhang2024stable,sun2024shmt}, we conduct a user study to complement quantitative evaluation. We randomly select 20 source-reference pairs from CPM-real~\cite{nguyen2021lipstick} and Wild-MT~\cite{jiang2020psgan}. 10 participants evaluated the generated images and selected the best one based on identity preservation, makeup consistency, non-edited region difference and image quality. The results reported in~\cref{tab:FRAM-user_study} show that our FRAM has the best performance in terms of human perception.

\setlength{\tabcolsep}{2.5pt}
\begin{table}[htb]
	\caption{\textbf{Ablation on CLIP fine-tuning objectives}. \textbf{SSL} is self-supervised contrastive learning. \textbf{Text} is image-text contrastive learning. \textbf{Acc} is K-nearest neighbour (KNN) classification accuracy on the makeup style data, where K is 5. The first row denotes off-the-shelf CLIP~\cite{radford2021learning} pre-trained for generic tasks.}
	\centering
	\begin{tabular}{c c c c c c c c}
		\toprule
		SSL & Text & Acc $\uparrow$ & CSD $\uparrow$ & ID $\uparrow$ & SSIM $\uparrow$ & L2-M $\downarrow$ & Aes $\uparrow$ \\
		\midrule
		\xmark & \xmark & 61.7 & 0.461 & \textbf{0.459} & \textbf{0.799} & 0.003 & 4.88 \\
		\checkmark & \xmark & 80.3 & 0.506 & 0.425 & 0.786 & 0.003 & 4.86 \\
		\xmark & \checkmark & 86.9 & 0.508 & 0.406 & 0.776 & 0.003 & 4.79 \\
		\checkmark & \checkmark & \textbf{88.2} & \textbf{0.528} & 0.429 & 0.797 & \textbf{0.003} & \textbf{4.95} \\
		\bottomrule
	\end{tabular}
	\label{tab:FRAM-ablation-CLIP}
\end{table}
\setlength{\tabcolsep}{6pt}

\setlength{\tabcolsep}{3pt}
\begin{table}[htb]
	\caption{\textbf{Ablation on identity and makeup injection modules}. \textbf{LoRA} is the LoRA fine-tuning of denoising model cross-attention layers for fusing the makeup features. \textbf{Pixel}/\textbf{3D} is the pixel/3D mesh guidance for the identity encoder, respectively.}
	\centering
	\begin{tabular}{c c c c c c c c}
		\toprule
		LoRA & Pixel & 3D & CSD $\uparrow$ & ID $\uparrow$ & SSIM $\uparrow$ & L2-M $\downarrow$ & Aes $\uparrow$ \\
		\midrule
		\checkmark & \checkmark & \xmark & 0.510 & 0.403 & 0.775 & 0.003 & 4.88 \\
		\checkmark & \xmark & \checkmark & \textbf{0.587} & 0.048 & 0.333 & 0.003 & \textbf{5.57} \\
		\xmark & \checkmark & \checkmark & 0.467 & \textbf{0.652} & \textbf{0.810} & 0.003 & 4.55 \\
		\checkmark & \checkmark & \checkmark & 0.528 & 0.429 & 0.797 & \textbf{0.003} & 4.95 \\
		\bottomrule
	\end{tabular}
	\label{tab:FRAM-ablation-inject}
\end{table}
\setlength{\tabcolsep}{6pt}

\setlength{\tabcolsep}{3pt}
\begin{table}[htb]
	\caption{\textbf{Ablation on attention loss}. \textbf{Diff} is the diffusion loss. \textbf{Attn} is the attention loss.}
	\centering
	\begin{tabular}{c c c c c c c}
		\toprule
		Diff & Attn & CSD $\uparrow$ & ID $\uparrow$ & SSIM $\uparrow$ & L2-M $\downarrow$ & Aes $\uparrow$ \\
		\midrule
		\checkmark & \xmark & 0.518 & 0.416 & \textbf{0.799} & 0.003 & 4.90 \\
		\checkmark & \checkmark & \textbf{0.528} & \textbf{0.429} & 0.797 & \textbf{0.003} & \textbf{4.95} \\
		\bottomrule
	\end{tabular}
	\label{tab:FRAM-ablation-inject_obj}
\end{table}
\setlength{\tabcolsep}{6pt}

\subsection{Ablation study}

\subsubsection{CLIP fine-tuning objectives}
In~\cref{tab:FRAM-ablation-CLIP}, we ablate the CLIP fine-tuning objectives. The proposed self-supervised and image-text contrastive learning help improve makeup consistency. Altogether, they result in the best makeup consistency and better balance between identity preservation and makeup consistency. These validate our CLIP fine-tuning for makeup encoding.

\subsubsection{Identity and makeup injection modules}
In~\cref{tab:FRAM-ablation-inject}, we evaluate the effectiveness of identity and makeup injection modules. We observe: \textbf{(1)} LoRA fine-tuning improves makeup consistency by fusing the CLIP makeup features. \textbf{(2)} Both the pixel-level information and facial structure of 3D mesh from source image contribute to the identity preservation. These validate our identity and makeup injection modules.

We also provide qualitative ablation in~\cref{fig:FRAM-ablation-module}, which is in line with the observations above. \cref{fig:FRAM-ablation-makeup} shows that without either of the learned CLIP makeup encoder or LoRA fine-tuning, the model struggles to capture the makeup style (\ie, exposed bare skin on the cheek). \cref{fig:FRAM-ablation-id} shows that pixel-level information helps preserve the appearance while 3D mesh helps with the facial structure. Altogether, they result in the best identity preservation.

\subsubsection{Attention loss}
In~\cref{tab:FRAM-ablation-inject_obj}, we study the effect of attention loss. The results show that the attention loss helps improve makeup consistency. More importantly, the visualization in~\cref{fig:FRAM-region} shows that the attention loss encourages the model to look at facial regions (\ie, skin, eyes and mouth) and thus enables region-specific makeup transfer.

\section{Conclusion}
In this work, we propose \textbf{F}acial \textbf{R}egion-\textbf{A}ware \textbf{M}akeup features (\textbf{FRAM}) for diffusion-based makeup transfer. It has two stages: \text{(1)} fine-tuning CLIP to learn a makeup encoder on synthesized makeup style data. \text{(2)} construct before-and-after makeup image pairs and use them to learn to inject identity and facial region-aware makeup features for makeup transfer. Our FRAM surpasses prior methods on global and region-specific makeup transfer tasks.


{
    \small
    \bibliographystyle{ieeenat_fullname}
    \bibliography{references}

@String{Computing = "Computing" }

@String{Computer = "{IEEE} Computer" }

@String{Springer = "Springer-Verlag" }

@inproceedings{li2018beautygan,
	author = {Li, Tingting and Qian, Ruihe and Dong, Chao and Liu, Si and Yan, Qiong and Zhu, Wenwu and Lin, Liang},
	title = {BeautyGAN: Instance-level Facial Makeup Transfer with Deep Generative Adversarial Network},
	year = {2018},
	isbn = {9781450356657},
	publisher = {Association for Computing Machinery},
	address = {New York, NY, USA},
	url = {https://doi.org/10.1145/3240508.3240618},
	doi = {10.1145/3240508.3240618},
	booktitle = {Proceedings of the 26th ACM International Conference on Multimedia},
	pages = {645–653},
	numpages = {9},
	location = {Seoul, Republic of Korea},
	series = {MM '18}
}

@inproceedings{gu2019ladn,
	author={Gu, Qiao and Wang, Guanzhi and Chiu, Mang Tik and Tai, Yu-Wing and Tang, Chi-Keung},
	booktitle={2019 IEEE/CVF International Conference on Computer Vision (ICCV)}, 
	title={LADN: Local Adversarial Disentangling Network for Facial Makeup and De-Makeup}, 
	year={2019},
	volume={},
	number={},
	pages={10480-10489},
	doi={10.1109/ICCV.2019.01058}
}

@inproceedings{jiang2020psgan,
	author={Jiang, Wentao and Liu, Si and Gao, Chen and Cao, Jie and He, Ran and Feng, Jiashi and Yan, Shuicheng},
	booktitle={2020 IEEE/CVF Conference on Computer Vision and Pattern Recognition (CVPR)}, 
	title={PSGAN: Pose and Expression Robust Spatial-Aware GAN for Customizable Makeup Transfer}, 
	year={2020},
	volume={},
	number={},
	pages={5193-5201},
	doi={10.1109/CVPR42600.2020.00524}
}

@inproceedings{nguyen2021lipstick,
	author={Nguyen, Thao and Tran, Anh Tuan and Hoai, Minh},
	booktitle={2021 IEEE/CVF Conference on Computer Vision and Pattern Recognition (CVPR)}, 
	title={Lipstick ain’t enough: Beyond Color Matching for In-the-Wild Makeup Transfer}, 
	year={2021},
	volume={},
	number={},
	pages={13300-13309},
	doi={10.1109/CVPR46437.2021.01310}
}

@inproceedings{deng2021spatially,
	author={Deng, Han and Han, Chu and Cai, Hongmin and Han, Guoqiang and He, Shengfeng},
	booktitle={2021 IEEE/CVF Conference on Computer Vision and Pattern Recognition (CVPR)}, 
	title={Spatially-invariant Style-codes Controlled Makeup Transfer}, 
	year={2021},
	volume={},
	number={},
	pages={6545-6553},
	doi={10.1109/CVPR46437.2021.00648}
}

@inproceedings{sun2022ssat,
	title={Ssat: A symmetric semantic-aware transformer network for makeup transfer and removal},
	author={Sun, Zhaoyang and Chen, Yaxiong and Xiong, Shengwu},
	booktitle={Proceedings of the AAAI Conference on artificial intelligence},
	pages={2325--2334},
	year={2022}
}

@inproceedings{xiang2022ramgan,
	author="Xiang, Jianfeng
	and Chen, Junliang
	and Liu, Wenshuang
	and Hou, Xianxu
	and Shen, Linlin",
	editor="Avidan, Shai
	and Brostow, Gabriel
	and Ciss{\'e}, Moustapha
	and Farinella, Giovanni Maria
	and Hassner, Tal",
	title="RamGAN: Region Attentive Morphing GAN for Region-Level Makeup Transfer",
	booktitle="Computer Vision -- ECCV 2022",
	year="2022",
	publisher="Springer Nature Switzerland",
	address="Cham",
	pages="719--735",
	isbn="978-3-031-20047-2"
}

@inproceedings{yan2023beautyrec,
	author={Yan, Qixin and Guo, Chunle and Zhao, Jixin and Dai, Yuekun and Loy, Chen Change and Li, Chongyi},
	booktitle={2023 IEEE/CVF Conference on Computer Vision and Pattern Recognition Workshops (CVPRW)}, 
	title={BeautyREC: Robust, Efficient, and Component-Specific Makeup Transfer}, 
	year={2023},
	volume={},
	number={},
	pages={1102-1110},
	doi={10.1109/CVPRW59228.2023.00117}
}

@inproceedings{sun2024content,
	author={Sun, Zhaoyang and Xiong, Shengwu and Chen, Yaxiong and Rong, Yi},
	booktitle={2024 IEEE/CVF Conference on Computer Vision and Pattern Recognition (CVPR)}, 
	title={Content-Style Decoupling for Unsupervised Makeup Transfer without Generating Pseudo Ground Truth}, 
	year={2024},
	volume={},
	number={},
	pages={7601-7610},
	doi={10.1109/CVPR52733.2024.00726}
}

@inproceedings{lu2025beautybank,
	author={Lu, Qianwen and Yang, Xingchao and Taketomi, Takafumi},
	booktitle={2025 IEEE/CVF Winter Conference on Applications of Computer Vision (WACV)}, 
	title={BeautyBank: Encoding Facial Makeup in Latent Space}, 
	year={2025},
	volume={},
	number={},
	pages={4183-4193},
	doi={10.1109/WACV61041.2025.00411}
}

@inproceedings{jin2024toward,
	author="Jin, Qiaoqiao
	and Chen, Xuanhong
	and Jin, Meiguang
	and Chen, Ying
	and Shi, Rui
	and Zheng, Yucheng
	and Zhu, Yupeng
	and Ni, Bingbing",
	editor="Leonardis, Ale{\v{s}}
	and Ricci, Elisa
	and Roth, Stefan
	and Russakovsky, Olga
	and Sattler, Torsten
	and Varol, G{\"u}l",
	title="Toward Tiny and High-Quality Facial Makeup with Data Amplify Learning",
	booktitle="Computer Vision -- ECCV 2024",
	year="2025",
	publisher="Springer Nature Switzerland",
	address="Cham",
	pages="340--356",
	isbn="978-3-031-73383-3"
}

@inproceedings{sun2024shmt,
	title={{SHMT}: Self-supervised Hierarchical Makeup Transfer via Latent Diffusion Models},
	author={Zhaoyang Sun and Shengwu Xiong and Yaxiong Chen and Fei Du and Weihua Chen and Fan Wang and Yi Rong},
	booktitle={The Thirty-eighth Annual Conference on Neural Information Processing Systems},
	year={2024},
	url={https://openreview.net/forum?id=EeXcOYf3Lg}
}

@inproceedings{ruan2025mad,
	title={MAD: Makeup All-in-One with Cross-Domain Diffusion Model},
	author={Ruan, Bo-Kai and Shuai, Hong-Han},
	booktitle={Proceedings of the Computer Vision and Pattern Recognition Conference},
	pages={749--758},
	year={2025}
}

@inproceedings{zhang2024stable,
	author = {Zhang, Yuxuan and Yuan, Yirui and Song, Yiren and Liu, Jiaming},
	title = {StableMakeup: When Real-World Makeup Transfer Meets Diffusion Model},
	year = {2025},
	isbn = {9798400715402},
	publisher = {Association for Computing Machinery},
	address = {New York, NY, USA},
	url = {https://doi.org/10.1145/3721238.3730702},
	doi = {10.1145/3721238.3730702},
	booktitle = {Proceedings of the Special Interest Group on Computer Graphics and Interactive Techniques Conference Conference Papers},
	articleno = {68},
	numpages = {9},
	location = {
	},
	series = {SIGGRAPH Conference Papers '25}
}

@article{sii2024gorgeous,
	title={Gorgeous: Create Your Desired Character Facial Makeup from Any Ideas},
	author={Sii, Jia Wei and Chan, Chee Seng},
	journal={arXiv preprint arXiv:2404.13944},
	year={2024}
}

@article{zhu2025flux,
	title={FLUX-Makeup: High-Fidelity, Identity-Consistent, and Robust Makeup Transfer via Diffusion Transformer},
	author={Zhu, Jian and Liu, Shanyuan and Li, Liuzhuozheng and Gong, Yue and Wang, He and Cheng, Bo and Ma, Yuhang and Wu, Liebucha and Wu, Xiaoyu and Leng, Dawei and others},
	journal={arXiv preprint arXiv:2508.05069},
	year={2025}
}

@article{yang2025ffhq,
	title={FFHQ-Makeup: Paired Synthetic Makeup Dataset with Facial Consistency Across Multiple Styles},
	author={Yang, Xingchao and Ueda, Shiori and Huang, Yuantian and Akiyama, Tomoya and Taketomi, Takafumi},
	journal={arXiv preprint arXiv:2508.03241},
	year={2025}
}

@article{wu2025evomakeup,
	title={EvoMakeup: High-Fidelity and Controllable Makeup Editing with MakeupQuad},
	author={Wu, Huadong and Fu, Yi and Li, Yunhao and Gao, Yuan and Du, Kang},
	journal={arXiv preprint arXiv:2508.05994},
	year={2025}
}

@inproceedings{somepalli2024measuring,
	title={Measuring Style Similarity in Diffusion Models},
	author={Somepalli, Gowthami and Gupta, Anubhav and Gupta, Kamal and Palta, Shramay and Goldblum, Micah and Geiping, Jonas and Shrivastava, Abhinav and Goldstein, Tom},
	booktitle="Computer Vision -- ECCV 2024",
	publisher="Springer Nature Switzerland",
	year={2024}
}

@inproceedings{ho2020denoising,
	author = {Ho, Jonathan and Jain, Ajay and Abbeel, Pieter},
	booktitle = {Advances in Neural Information Processing Systems},
	editor = {H. Larochelle and M. Ranzato and R. Hadsell and M.F. Balcan and H. Lin},
	pages = {6840--6851},
	publisher = {Curran Associates, Inc.},
	title = {Denoising Diffusion Probabilistic Models},
	url = {https://proceedings.neurips.cc/paper_files/paper/2020/file/4c5bcfec8584af0d967f1ab10179ca4b-Paper.pdf},
	volume = {33},
	year = {2020}
}

@inproceedings{esser2021taming,
	author={Esser, Patrick and Rombach, Robin and Ommer, Björn},
	booktitle={2021 IEEE/CVF Conference on Computer Vision and Pattern Recognition (CVPR)}, 
	title={Taming Transformers for High-Resolution Image Synthesis}, 
	year={2021},
	volume={},
	number={},
	pages={12868-12878},
	doi={10.1109/CVPR46437.2021.01268}
}

@inproceedings{rombach2022high,
	author={Rombach, Robin and Blattmann, Andreas and Lorenz, Dominik and Esser, Patrick and Ommer, Björn},
	booktitle={2022 IEEE/CVF Conference on Computer Vision and Pattern Recognition (CVPR)}, 
	title={High-Resolution Image Synthesis with Latent Diffusion Models}, 
	year={2022},
	volume={},
	number={},
	pages={10674-10685},
	doi={10.1109/CVPR52688.2022.01042}
}

@inproceedings{zhang2023adding,
	title={Adding conditional control to text-to-image diffusion models},
	author={Zhang, Lvmin and Rao, Anyi and Agrawala, Maneesh},
	booktitle={Proceedings of the IEEE/CVF International Conference on Computer Vision},
	pages={3836--3847},
	year={2023}
}

@inproceedings{gal2023an,
	title={An Image is Worth One Word: Personalizing Text-to-Image Generation using Textual Inversion},
	author={Rinon Gal and Yuval Alaluf and Yuval Atzmon and Or Patashnik and Amit Haim Bermano and Gal Chechik and Daniel Cohen-or},
	booktitle={The Eleventh International Conference on Learning Representations },
	year={2023},
	url={https://openreview.net/forum?id=NAQvF08TcyG}
}

@inproceedings{radford2021learning,
	title = 	 {Learning Transferable Visual Models From Natural Language Supervision},
	author =       {Radford, Alec and Kim, Jong Wook and Hallacy, Chris and Ramesh, Aditya and Goh, Gabriel and Agarwal, Sandhini and Sastry, Girish and Askell, Amanda and Mishkin, Pamela and Clark, Jack and Krueger, Gretchen and Sutskever, Ilya},
	booktitle = 	 {Proceedings of the 38th International Conference on Machine Learning},
	pages = 	 {8748--8763},
	year = 	 {2021},
	editor = 	 {Meila, Marina and Zhang, Tong},
	volume = 	 {139},
	series = 	 {Proceedings of Machine Learning Research},
	month = 	 {18--24 Jul},
	publisher =    {PMLR},
	url = 	 {https://proceedings.mlr.press/v139/radford21a.html},
}

@article{labs2025flux,
	title={FLUX. 1 Kontext: Flow Matching for In-Context Image Generation and Editing in Latent Space},
	author={Labs, Black Forest and Batifol, Stephen and Blattmann, Andreas and Boesel, Frederic and Consul, Saksham and Diagne, Cyril and Dockhorn, Tim and English, Jack and English, Zion and Esser, Patrick and others},
	journal={arXiv preprint arXiv:2506.15742},
	year={2025}
}

@inproceedings{hu2022lora,
	title={Lo{RA}: Low-Rank Adaptation of Large Language Models},
	author={Edward J Hu and yelong shen and Phillip Wallis and Zeyuan Allen-Zhu and Yuanzhi Li and Shean Wang and Lu Wang and Weizhu Chen},
	booktitle={International Conference on Learning Representations},
	year={2022},
	url={https://openreview.net/forum?id=nZeVKeeFYf9}
}

@inproceedings{zheng2022general,
	author={Zheng, Yinglin and Yang, Hao and Zhang, Ting and Bao, Jianmin and Chen, Dongdong and Huang, Yangyu and Yuan, Lu and Chen, Dong and Zeng, Ming and Wen, Fang},
	booktitle={2022 IEEE/CVF Conference on Computer Vision and Pattern Recognition (CVPR)}, 
	title={General Facial Representation Learning in a Visual-Linguistic Manner}, 
	year={2022},
	volume={},
	number={},
	pages={18676-18688},
	doi={10.1109/CVPR52688.2022.01814}
}

@book{wahba1990spline,
	title={Spline models for observational data},
	author={Wahba, Grace},
	year={1990},
	publisher={SIAM}
}

@inproceedings{pmlr-v119-chen20j,
	title = {A Simple Framework for Contrastive Learning of Visual Representations},
	author = {Chen, Ting and Kornblith, Simon and Norouzi, Mohammad and Hinton, Geoffrey},
	booktitle = {Proceedings of the 37th International Conference on Machine Learning},
	pages = {1597--1607},
	year = {2020},
	editor = {Hal Daumé III and Aarti Singh},
	volume = {119},
	series = {Proceedings of Machine Learning Research},
	month = {13--18 Jul},
	publisher = {PMLR},
	url = { http://proceedings.mlr.press/v119/chen20j.html }
}

@inproceedings{khosla2020supervised,
	author = {Khosla, Prannay and Teterwak, Piotr and Wang, Chen and Sarna, Aaron and Tian, Yonglong and Isola, Phillip and Maschinot, Aaron and Liu, Ce and Krishnan, Dilip},
	booktitle = {Advances in Neural Information Processing Systems},
	editor = {H. Larochelle and M. Ranzato and R. Hadsell and M.F. Balcan and H. Lin},
	pages = {18661--18673},
	publisher = {Curran Associates, Inc.},
	title = {Supervised Contrastive Learning},
	url = {https://proceedings.neurips.cc/paper/2020/file/d89a66c7c80a29b1bdbab0f2a1a94af8-Paper.pdf},
	volume = {33},
	year = {2020}
}

@inproceedings{carion2020end,
	author="Carion, Nicolas
	and Massa, Francisco
	and Synnaeve, Gabriel
	and Usunier, Nicolas
	and Kirillov, Alexander
	and Zagoruyko, Sergey",
	editor="Vedaldi, Andrea
	and Bischof, Horst
	and Brox, Thomas
	and Frahm, Jan-Michael",
	title="End-to-End Object Detection with Transformers",
	booktitle="Computer Vision -- ECCV 2020",
	year="2020",
	publisher="Springer International Publishing",
	address="Cham",
	pages="213--229",
	isbn="978-3-030-58452-8"
}

@inproceedings{lin2017focal,
	author={Lin, Tsung-Yi and Goyal, Priya and Girshick, Ross and He, Kaiming and Dollár, Piotr},
	booktitle={2017 IEEE International Conference on Computer Vision (ICCV)}, 
	title={Focal Loss for Dense Object Detection}, 
	year={2017},
	volume={},
	number={},
	pages={2999-3007},
	doi={10.1109/ICCV.2017.324}
}

@inproceedings{milletari2016v,
	author={Milletari, Fausto and Navab, Nassir and Ahmadi, Seyed-Ahmad},
	booktitle={2016 Fourth International Conference on 3D Vision (3DV)}, 
	title={V-Net: Fully Convolutional Neural Networks for Volumetric Medical Image Segmentation}, 
	year={2016},
	volume={},
	number={},
	pages={565-571},
	doi={10.1109/3DV.2016.79}
}

@inproceedings{guo2022perspective,
	author="Guo, Jia
	and Yu, Jinke
	and Lattas, Alexandros
	and Deng, Jiankang",
	editor="Karlinsky, Leonid
	and Michaeli, Tomer
	and Nishino, Ko",
	title="Perspective Reconstruction of Human Faces by Joint Mesh and Landmark Regression",
	booktitle="Computer Vision -- ECCV 2022 Workshops",
	year="2023",
	publisher="Springer Nature Switzerland",
	address="Cham",
	pages="350--365",
	isbn="978-3-031-25072-9"
}

@inproceedings{goodfellow2014generative,
	author = {Goodfellow, Ian J. and Pouget-Abadie, Jean and Mirza, Mehdi and Xu, Bing and Warde-Farley, David and Ozair, Sherjil and Courville, Aaron and Bengio, Yoshua},
	booktitle = {Advances in Neural Information Processing Systems},
	editor = {Z. Ghahramani and M. Welling and C. Cortes and N. Lawrence and K.Q. Weinberger},
	pages = {},
	publisher = {Curran Associates, Inc.},
	title = {Generative Adversarial Nets},
	url = {https://proceedings.neurips.cc/paper_files/paper/2014/file/f033ed80deb0234979a61f95710dbe25-Paper.pdf},
	volume = {27},
	year = {2014}
}

@article{wang2024instantid,
	title={Instantid: Zero-shot identity-preserving generation in seconds},
	author={Wang, Qixun and Bai, Xu and Wang, Haofan and Qin, Zekui and Chen, Anthony and Li, Huaxia and Tang, Xu and Hu, Yao},
	journal={arXiv preprint arXiv:2401.07519},
	year={2024}
}

@article{ye2023ip,
	title={Ip-adapter: Text compatible image prompt adapter for text-to-image diffusion models},
	author={Ye, Hu and Zhang, Jun and Liu, Sibo and Han, Xiao and Yang, Wei},
	journal={arXiv preprint arXiv:2308.06721},
	year={2023}
}

@inproceedings{gao2024self,
	author={Gao, Zheng and Patras, Ioannis},
	booktitle={2024 IEEE/CVF Conference on Computer Vision and Pattern Recognition (CVPR)}, 
	title={Self-Supervised Facial Representation Learning with Facial Region Awareness}, 
	year={2024},
	volume={},
	number={},
	pages={2081-2092},
	doi={10.1109/CVPR52733.2024.00203}
}

@inproceedings{dosovitskiy2021an,
	title={An Image is Worth 16x16 Words: Transformers for Image Recognition at Scale},
	author={Alexey Dosovitskiy and Lucas Beyer and Alexander Kolesnikov and Dirk Weissenborn and Xiaohua Zhai and Thomas Unterthiner and Mostafa Dehghani and Matthias Minderer and Georg Heigold and Sylvain Gelly and Jakob Uszkoreit and Neil Houlsby},
	booktitle={International Conference on Learning Representations},
	year={2021},
	url={https://openreview.net/forum?id=YicbFdNTTy}
}

@misc{ControlNet++,
	author       = {xinsir6},
	title        = {ControlNet++: All-in-one ControlNet for image generations and editing},
	year         = {2024},
	howpublished = {},
	note         = {Accessed: 2025-11-08}
}

@inproceedings{wang20243d,
	author={Wang, Zidu and Zhu, Xiangyu and Zhang, Tianshuo and Wang, Baiqin and Lei, Zhen},
	booktitle={2024 IEEE/CVF Conference on Computer Vision and Pattern Recognition (CVPR)}, 
	title={3D Face Reconstruction with the Geometric Guidance of Facial Part Segmentation}, 
	year={2024},
	volume={},
	number={},
	pages={1672-1682},
	doi={10.1109/CVPR52733.2024.00165}
}

@inproceedings{karras2019style,
	author={Karras, Tero and Laine, Samuli and Aila, Timo},
	booktitle={2019 IEEE/CVF Conference on Computer Vision and Pattern Recognition (CVPR)}, 
	title={A Style-Based Generator Architecture for Generative Adversarial Networks}, 
	year={2019},
	volume={},
	number={},
	pages={4396-4405},
	doi={10.1109/CVPR.2019.00453}
}

@inproceedings{shiohara2023blendface,
	author={Shiohara, Kaede and Yang, Xingchao and Taketomi, Takafumi},
	booktitle={2023 IEEE/CVF International Conference on Computer Vision (ICCV)}, 
	title={BlendFace: Re-designing Identity Encoders for Face-Swapping}, 
	year={2023},
	volume={},
	number={},
	pages={7600-7610},
	doi={10.1109/ICCV51070.2023.00702}
}

@article{wang2004image,
	author={Zhou Wang and Bovik, A.C. and Sheikh, H.R. and Simoncelli, E.P.},
	journal={IEEE Transactions on Image Processing}, 
	title={Image quality assessment: from error visibility to structural similarity}, 
	year={2004},
	volume={13},
	number={4},
	pages={600-612},
	doi={10.1109/TIP.2003.819861}
}

@inproceedings{schuhmann2022laionb,
	title={{LAION}-5B: An open large-scale dataset for training next generation image-text models},
	author={Christoph Schuhmann and Romain Beaumont and Richard Vencu and Cade W Gordon and Ross Wightman and Mehdi Cherti and Theo Coombes and Aarush Katta and Clayton Mullis and Mitchell Wortsman and Patrick Schramowski and Srivatsa R Kundurthy and Katherine Crowson and Ludwig Schmidt and Robert Kaczmarczyk and Jenia Jitsev},
	booktitle={Thirty-sixth Conference on Neural Information Processing Systems Datasets and Benchmarks Track},
	year={2022},
	url={https://openreview.net/forum?id=M3Y74vmsMcY}
}

@inproceedings{feng2026noisy,
	title={Noisy but Valid: Robust Statistical Evaluation of {LLM}s with Imperfect Judges},
	author={Chen Feng and Minghe Shen and Ananth Balashankar and Carsten Gerner-Beuerle and Miguel R. D. Rodrigues},
	booktitle={The Fourteenth International Conference on Learning Representations},
	year={2026},
	url={https://openreview.net/forum?id=hEhxreaLdU}
}

@inproceedings{meng2024mm2latent,
	author="Meng, Debin
	and Tzelepis, Christos
	and Patras, Ioannis
	and Tzimiropoulos, Georgios",
	editor="Del Bue, Alessio
	and Canton, Cristian
	and Pont-Tuset, Jordi
	and Tommasi, Tatiana",
	title="MM2Latent: Text-to-Facial Image Generation and Editing in GANs with Multimodal Assistance",
	booktitle="Computer Vision -- ECCV 2024 Workshops",
	year="2025",
	publisher="Springer Nature Switzerland",
	address="Cham",
	pages="88--106",
	isbn="978-3-031-91838-4"
}

@article{meng2025training,
	title={Training-Free Generation of Diverse and High-Fidelity Images via Prompt Semantic Space Optimization},
	author={Meng, Debin and Jin, Chen and Gao, Zheng and Li, Yanran and Patras, Ioannis and Tzimiropoulos, Georgios},
	journal={arXiv preprint arXiv:2511.19811},
	year={2025}
}
}


\end{document}